\documentclass[10pt, a4paper]{article}
\usepackage{lrec-coling2024} % this is the new style

\usepackage{natbib}
\usepackage{multibib}
\makeatletter
\def\@mb@citenamelist{cite,citep,citet,citealp,citealt,citepalias,citetalias}
\makeatother
\newcites{languageresource}{~}

\usepackage{graphicx}
\usepackage{tabularx}
\usepackage{soul}
\usepackage{siunitx}
\usepackage{titlesec}
\titleformat{\section}{\normalfont\large\bfseries\center}{\thesection.}{1em}{}
\titleformat{\subsection}{\normalfont\SmallTitleFont\bfseries\raggedright}{\thesubsection.}{1em}{}
\titleformat{\subsubsection}{\normalfont\normalsize\bfseries\raggedright}{\thesubsubsection.}{1em}{}
\renewcommand\thesection{\arabic{section}}
\renewcommand\thesubsection{\thesection.\arabic{subsection}}
\renewcommand\thesubsubsection{\thesubsection.\arabic{subsubsection}}
\usepackage{xcolor}
\usepackage{hyperref}
 \definecolor{darkblue}{rgb}{0, 0, 0.5}
  \hypersetup{colorlinks=true, citecolor=darkblue, linkcolor=darkblue, urlcolor=darkblue}

\usepackage{xstring}

\usepackage{color}

\title{Solving Word-Sense Disambiguation and Word-Sense Induction with Dictionary Examples}

\name{Tadej Škvorc and Marko Robnik-Šikonja} 

\address{University of Ljubljana, Faculty of Computer and Information Science\\
Večna pot 113, 1000 Ljubljana, Slovenia \\
tadej.skvorc@fri.uni-lj.si,marko.robnik@fri.uni-lj.si}

\abstract{
Many less-resourced languages struggle with a lack of large, task-specific datasets that are required for solving relevant tasks with modern transformer-based large language models (LLMs). On the other hand, many linguistic resources, such as dictionaries, are rarely used in this context despite their large information contents. We show how LLMs can be used to extend existing language resources in less-resourced languages for two important tasks: word-sense disambiguation (WSD) and word-sense induction (WSI). We approach the two tasks through the related but much more accessible word-in-context (WiC) task where, given a pair of sentences and a target word, a classification model is tasked with predicting whether the sense of a given word differs between sentences. We demonstrate that a well-trained model for this task can distinguish between different word senses and can be adapted to solve the WSD and WSI tasks. The advantage of using the WiC task, instead of directly predicting senses, is that the WiC task does not need pre-constructed sense inventories with a sufficient number of examples for each sense, which are rarely available in less-resourced languages. We show that sentence pairs for the WiC task can be successfully generated from dictionary examples using LLMs. The resulting prediction models outperform existing models on WiC, WSD, and WSI tasks. We demonstrate our methodology on the Slovene language, where a monolingual dictionary is available, but word-sense resources are tiny.
 \\ \newline  }

\begin{document}

\maketitleabstract

\section{Introduction}
State-of-the-art approaches in most natural language processing (NLP) tasks such as machine translation \cite{dabre2020survey}, text generation \cite{iqbal2022survey}, and word-sense disambiguation \cite{bevilacqua2021recent} currently rely on pre-trained large language models (LLMs). In spite of LLMs' versatility, for many complex tasks, we still require relatively large, labeled datasets to fine-tune the models. While such datasets are available in English and other well-resourced languages, they rarely exist in the majority of other, less-resourced languages. Consequently, NLP tools in such languages often support significantly fewer tasks compared to English. Finding alternative approaches to create the necessary data is therefore a research of high significance and practical relevance.

One way to address the lack of data is to automatically generate training instances using LLMs. A key advantage of pre-trained LLMs is their self-supervised training, i.e. that they can be trained on unlabelled text \cite{shorten2021text}. Such generative approaches are usually task-specific and generate new data based on existing examples or domain knowledge. % for example, by modifying existing images to produce new images for tasks in computer vision \cite{shorten2019survey}. 
For text, this was traditionally done with rule-based approaches and, recently, through auxiliary LLMs to augment existing data. Such models take a (weakly related) text and extend it in a way that tries to mimic human-generated language. We argue that traditional linguistic resources, such as dictionaries, available in many less-resourced languages, can be a source of relevant texts and information. 

Recently, generative LLMs have advanced significantly and models such as LaMDA \cite{thoppilan2022lamda}, GPT-3 \cite{brown2020language}, GPT-4 \cite{openai2023gpt4} and Llama-3 \cite{dubey2024llama3herdmodels} show excellent performance in many tasks. In comparison to traditional language models, LLMs are trained on larger datasets, contain larger amounts of network parameters, and are human-aligned with instruction fine-tuning, allowing them to generate texts that closely match those produced by humans. Furthermore, existing work has shown that such models can be used to solve specific tasks without the need for manually annotated data with the so-called in-context learning. Examples include solving grade-school math assignments \cite{cobbe2021training} and generating code in a variety of programming languages \cite{chen2021evaluating}. This makes them potentially useful for a variety of tasks that currently suffer from a lack of large annotated datasets.

\begin{figure*}[!htb]
\begin{center}
%\fbox{\parbox{6cm}{
%This is a figure with a caption.}}
% old picture \includegraphics[scale=0.5]{lrec2020W-image1.eps} 
%\includegraphics[trim=120 100 120 100,clip,width=\textwidth]{flowchartr_2.jpg} 
\includegraphics[trim=150 100 100 100,clip,width=\textwidth]{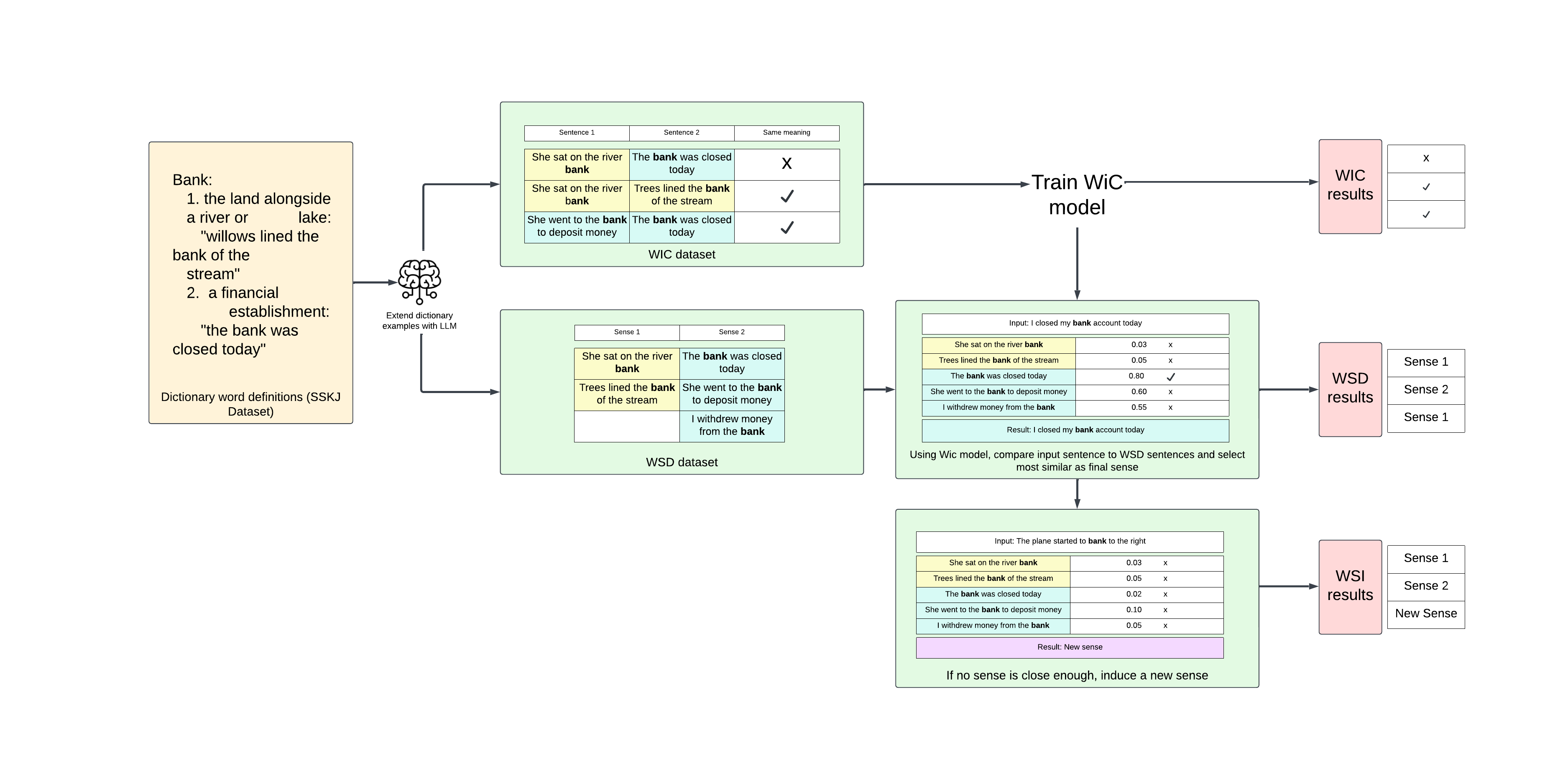} 

\caption{A flowchart demonstrating our methodology of using WiC for WSD \& WSI. Dictionary definitions and usage snippets can be extended into WiC dataset that is used to create a same-sense classifier. This WiC classifier can be applied to all different word senses to solve the WSD task. If no sense is matched, we have a candidate for a new sense, i.e. addressing the WSI task. The WSD dataset can be replaced with examples generated from a dictionary for less-resourced languages.  }
\label{fig:flowchart}
\end{center}
\end{figure*}

In the word-sense disambiguation (WSD) task, a model tries to identify the correct meaning of a word in a given context, especially when the word has multiple possible meanings or senses. This task relies on a sense inventory and is crucial for understanding human language, as many words are polysemous and their correct interpretation depends on the specific context in which they are used. WSD is important for applications such as machine translation, information retrieval, and (linguistic) text analyses. Unlike WSD, which maps words to known senses from a dictionary or database, word-sense induction (WSI) discovers and clusters possible senses or meanings directly from text corpora. This enables handling new word senses in languages with sense inventories and is useful in languages or domains where sense-annotated resources are not available.

Most existing research in NLP focuses on English and most LLMs are primarily trained on English texts. However, as several LLMs intentionally or unintentionally contain small amounts of training data in other languages, their performance on these languages currently remains underexplored. In our work, we examine whether LLMs could be helpful in improving NLP tasks in a less-resourced language based on commonly available linguistic resources. Specifically, we examine if GPT-3.5 can augment and extend training data for Slovene word-sense disambiguation (WSD) and word-sense induction (WSI) based on dictionary entries.
 We start with a dataset of dictionary entries, where each word sense contains examples that are too short and incomplete for training WSD models using contextual models, such as BERT \cite{Devlin2019}. We show that LLMs can be used to extend these short dictionary examples into complete sentences while maintaining a correct sense of the given word. 

An additional problem with models for the WSD task in this setting is that they are difficult to evaluate, as WSD models traditionally require a unified sense inventory in order to match words to senses. Such inventories are unavailable in many languages or may not match well the senses captured by LLMs. To address this, we use an alternative Word-in-Context (WiC) classification task presented by \citet{martelli-etal-2021-semeval}. In this task, a model is presented with a pair of sentences containing the same target lemma and has to determine whether its sense differs between the two sentences. We show that if data generated by LLM is used in the WiC task, the resulting models exhibit an improved classification performance compared to models that do not use such data. Finally, we show that models trained on the WiC tasks can be adapted to solve WSD and WSI tasks through pairwise comparison with sense-annotated data.
An outline of our approach is depicted in Figure \ref{fig:flowchart}.

Our main contributions are as follows: 
\begin{enumerate}
\item  A novel methodology to utilize information from traditional language resources (such as dictionaries) to create datasets for the WiC task using LLMs. \\
\item A scheme that enables models trained on the WiC task to be used for WSD and WSI tasks. \\
\item A demonstration of the proposed methodologies on the Slovene language where WiC, WSD, and WSI datasets are practically non-existent. Our results show that the proposed pipeline produces practically useful WSD and WSI prediction models.
\end{enumerate}

The paper is structured into five further sections. Section \ref{sec:related} presents an overview of related work. In Section \ref{sec:methodology}, we describe the WiC task and its relation to the more common WSD and WSI tasks and show how models trained on WiC tasks can be used for WSI and WSD tasks. We also describe how we use an LLM to extend dictionary definitions and examples into full sentences, preserving their meaning.  In Section \ref{sec:dataset}, we describe our training datasets for the WiC task, provide a qualitative and quantitative evaluation of LLM-generated sentences and describe how the WiC dataset can be used to construct datasets targeting WSD and WSI tasks. Section \ref{sec:results} contains the results and their interpretation. We conclude with a discussion and ideas for further work in Section \ref{sec:conclusions}. 

\section{Related work}
\label{sec:related}
Traditionally, WSD tasks can be tackled by first constructing a large dataset where each word is manually labeled with its corresponding sense and then using this dataset to train a classification model. Early approaches involve classical machine learning (ML) models, such as support vector machines \cite{lee2004supervised} and naive Bayes classification \cite{escudero2000naive}. Another set of early approaches makes use of dictionaries and other linguistic resources (e.g., WordNet \cite{miller1995wordnet}), which contain information related to word senses and can be used for WSD \cite{li1995wordnet,banerjee2002adapted}. In contrast, WSI is commonly tackled using unsupervised methods, conceptualizing the task as a clustering problem and using the resulting clusters to assign senses \cite{navigli2012quick, denkowski2009survey}. 

%TODO cite SloWnet
More recently, deep neural networks have become the most common approach for WSD \cite{bevilacqua2021recent,loureiro2020language}. Deep learning approaches commonly outperform simpler machine learning models but can often require larger training datasets. While such approaches are suitable for well-resourced languages such as English or German, they are less suitable for less-resourced languages where large, manually-labeled datasets do not exist. For example, Slovene contains a limited amount of WSD resources. Its WSD approaches often make use of SloWNet, a Slovene translation of WordNet, or rely on multilingual datasets such as Elexis-WSD \cite{martelli2021designing}, making use of cross-lingual transfer \cite{fijavvz2023slovene}. This makes methods that rely on large amounts of training data (e.g., deep neural networks) difficult to use and leads to lower model performance compared to well-resourced languages.

One way to overcome this drawback is by using closely related tasks that do not rely on intensive manual annotation. \citet{martelli-etal-2021-semeval} present the word-in-context task, which presents a model with a pair of sentences and a target lemma and requires the model to predict whether the lemma's sense matches or differs between sentences. Such datasets are more frequently available in less-resourced languages, e.g., in Slovene, such a dataset was presented by \citet{knez2023word}.

Another approach involves generating the training data automatically, thus reducing the need for time-consuming manual annotation, either by augmenting existing examples (data augmentation) or by creating new examples from scratch (data generation). In recent years, large language models have been shown to generate datasets that can be useful in a variety of tasks, including text mining \cite{tang2023does}, text classification \cite{li2023synthetic}, and stance detection \cite{wagner2024power}. Generating increasingly human-like texts makes LMMs well-suited for WSD data generation. However, care must be taken that the senses in the generated data match their annotations. \citet{janz2023wordnet} present an approach that generated data using a model trained on SemCor and WordNet datasets. \citet{saidi2022gpt} present a similar approach using a GPT-2 LLM fine-tuned on the SemCor dataset.

Alternatively, LLMs can be used to extend existing data from various language resources (e.g., dictionaries) without the need for fine-tuning on WSD datasets. \citet{cai2024low} provide an LLM with a word, its dictionary definition, and its part of speech and show that LMMs can generate high-quality example sentences that correctly match the given word and sense.

Our work combines dictionary-based LLM data generation with the WiC task, which is simpler than WSD, thus reducing the need for manually constructed datasets when training Slovene WSD models.

\section{Using Dictionaries for WiC, WSD and WSI}
\label{sec:methodology}
A standard approach to tackle the WSD task, and to a lesser extent also the WSI task, requires a unified sense inventory where each word sense is accompanied by a sufficiently large set of sentences where the word is used in a given sense. Such inventories are unavailable in many languages.
 
We avoid this problem by building a prediction model for a simpler word-in-context (WiC) task. The binary WiC classifier is presented with a pair of sentences containing the same target lemma and determines if its sense differs between the two sentences. Table \ref{tab:wic_example} shows two example pairs:

\begin{table}[htb]
\caption{\label{tab:wic_example} Example sentence pairs for the WiC tasks. The classifier returns 1 when the sense of the target word (indicated by [t] ) is the same in both sentences and 0 if it is not.
}

\begin{tabular}{l}
Sentence\\
\hline
The [t] bank was closed on Sunday.\\
She sat on the river [t] bank.\\

Class: 0 \\
\hline
She went to the [t] bank to deposit money. \\
The [t] bank was closed on Sunday. \\
Class: 1\\

%Total sentences & 200 & 100\%\\
\hline
\end{tabular}
\end{table}

The WiC task does not directly assign senses like the WSD task does, removing the need for a predetermined sense inventory. However, we can use the WiC task to assign senses by repeatedly running the WiC classifier, taking as an input a given sentence paired with sentence examples for all different senses of the given word. A sense with the most certain match can be selected as the final output. 

For the WSI task, which searches for new word senses, the procedure of applying the WiC classifier to examples of all word senses remains the same. However, if none of the existing senses is reliably matched to a word sense in the given sentence, this word sense becomes a candidate for a new sense. 

We show the diagram of our complete process, including dataset generation and WiC, WSD, and WSI tasks in Figure \ref{fig:flowchart}. We start by creating a dataset for the WiC task. In Section \ref{sec:dictionaryDefs}, we describe the structure of a dictionary used to create the dataset. In Section \ref{sec:LLM4expansion}, we describe how we create the dataset by applying an LLM to dictionary sense definitions and very short examples as seed information. As adequate monolingual or bilingual dictionaries with such definitions and examples exist for many languages, the proposed methodology is widely applicable.

\subsection{Dictionaries as WiC Resources}
\label{sec:dictionaryDefs}
To form a dataset for WSD, WSI, or WiC classification, we need complete sentences covering different word senses. For each lemma, most dictionaries provide definitions of word senses and, for each sense, a short example outlining its use (see Table \ref{tab:sskj}). Neither definitions nor short examples are adequate for the WSD, WSI or WiC tasks. However, we show that 
%We show that data generated by LLM can be used to improve classification performance over models that do not use such data.
LLMs, such as GPT-3.5, can augment and extend this data into complete sentences suitable for training data for Slovene WSD, WSI, and WiC. We first describe our dictionary resource, followed by the expansion of sense definitions and examples into complete sentences with GPT-3.5.

\subsubsection*{Dictionary structure}
\label{sec:dictionary_structure}
In a typical dictionary, each entry contains examples of word senses that are too short and incomplete for training WSD classifiers using context-sensitive models, such as BERT \cite{Devlin2019} (see an example in Table \ref{tab:sskj}). We use GPT-3.5 to extend these short dictionary examples into complete sentences while maintaining the correct sense of a given word. 

We extend word descriptions from the online version of the Dictionary of Standard Slovene Language (Slovar slovenskega knjižnega jezika, SSKJ) \cite{SSKJ}. SSKJ contains 91,318 dictionary entries, i.e. Slovene words, where each entry contains the following data relevant to our purpose:
    \begin{itemize}
        \item the word lemma,
        \item definitions of all word senses,
        \item short usage examples for each word sense,
        \item phrases, expressions, and slang connected to the word.
    \end{itemize}

The SSKJ entry for the word \emph{slovar} (in English: dictionary) in Table \ref{tab:sskj} contains two senses. Most commonly, the word is used to describe a book containing word definitions. Alternatively, it can be used to describe a person's vocabulary. 

\begin{table*}
\centering
\caption{\label{tab:sskj}An example of an entry in the Slovene SSKJ dictionary, with the original Slovene text on the left and its English translation on the right. The word shown is \emph{slovar} (in English: dictionary).
}

\begin{tabular}{p{0.5\linewidth}p{0.5\linewidth}}
& \\
\hline
1. knjiga, v kateri so besede razvrščene po abecedi in pojasnjene: slovar ima sto tisoč besed; izdati, sestavljati slovar; prevajati s slovarji; obsežen slovar / na koncu knjige je slovar seznam s tako razvrščenimi in pojasnjenimi besedami / enojezični, narečni, pravopisni, tehniški slovar; [...]

* jezikosl. avtorski slovar ki vsebuje besede določenega avtorja; etimološki, frekvenčni, informativno-normativni slovar; obrnjeni slovar; [...]

2. besedni zaklad: imeti bogat slovar / njen slovar ni bil ravno izbran
** ekspr. besede nemogoče ni v njegovem slovarju\textit{ odločen je narediti tudi na videz nemogoče stvari}; ekspr. če to povemo v ekonomskem slovarju [...]

&

1. a book containing alphabetically-sorted words with explanations: the dictionary contains one hundred thousand words; to publish, create a dictionary; to translate using dictionaries; a comprehensive dictionary / the end of the book contains a dictionary with sorted words and definitions / monolingual, dialect, spelling, technical dictionary [...]

* ling. author vocabulary with words of certain author; etymological, frequency, informative-normative; reverse dictionary; [...]

2. a vocabulary: to have a large vocabulary / her vocabulary was lacking
** expr. the word impossible is not in his dictionary \textit{he is determined to do things that seem impossible}; expr. if we use a word from an economic dictionary [...] \\
\hline
\end{tabular}

\end{table*}

Dictionary definitions in SSKJ are written as plain text whose structure is easy to parse. Every word sense begins with a sequential number followed by the definition that ends with a colon. Afterward, short use-case examples are listed, separated by semicolons or forward slashes for related examples. Additionally, each sense contains specialized examples of various types (phrases, slang, expressive languages, terminology, etc.), denoted by a special symbol (* or **) and the type (e.g., expr. for expressive language or ling. for linguistic terminology).

\subsection{LLM for sentence expansion}
\label{sec:LLM4expansion}
While the information contained in dictionaries can be of interest for various NLP tasks, we focus on word senses. The examples of usage for each word sense illustrate a typical use of the given word sense.  We extract all usage examples, discarding other text. The selected examples are usually only a few words long and almost never form complete sentences, making them poorly suited for training WSD models. While these sentence snippets do contain some semantic information, they are unlikely to be similar to sentences encountered in texts on which LLMs are pre-trained. While it is possible to train WSD models on these short usage examples, we later show that the short length makes them less suitable than a dataset containing complete sentences.

Manually creating a dataset with complete sentences would be time- and cost-prohibitive. Instead, we generate expanded sentences automatically using a generative LLM, with the usage examples given as starting information. In our work, we use GPT-3.5 (text-davinci-003) model. LLMs are ideal for this task because their large size and huge amounts of training data allow them to generate text that closely matches human language.

Many capabilities of LLMs in less-resourced languages, such as Slovene, are currently underexplored. We analyze if GPT-3.5 can generate useful sentences for WSD in Slovene based on dictionary usage examples. There are several possible failures. First, it is possible that GPT-3.5 can only generate words from a small number of senses, i.e. not for all senses available in a dictionary. Second, and more importantly, we do not know if GPT-3.5 is able to take a short dictionary definition and expand it into a full sentence so that the target word matches the sense in the original definition.

For our task, it is especially important that the sentences produced by the LLM use the word in the same sense as the original usage example. To evaluate whether this is the case, we generate complete sentences using the following prompt: "Razširi [zgled] v polno poved" (Expand [usage example] into a full sentence), where [zgled] ([usage example]) is the usage example obtained from the SSKJ dictionary. Since the usage example already contains words related to the correct word sense, this should guide the LLM to generate complete sentences with the correct word sense. To validate this assumption, we generated one complete sentence for each dictionary usage example and examined the results using manual and automatic methods. The procedure and results of this evaluation are presented in Section \ref{sec:results}.

While we could apply our approach to every word in the SSKJ dictionary,  we limited our evaluation to the words that appear in the Elexis Parallel Sense-Annotated dataset \cite{martelli2021designing}, as this is currently the largest Slovene WSD dataset. We provide a description of this dataset in Section \ref{sec:elexis-wsd}.

\section{Datasets}
\label{sec:dataset}
In order to evaluate if our approach is applicable to the WSD and WSI tasks, we use three datasets. The first is SSKJ-WSD\footnote{\url{https://www.clarin.si/repository/xmlui/handle/11356/2008}}, a newly constructed dataset that consists of SSKJ dictionary usage examples converted to complete sentences using GPT-3.5. This dataset, described in Section \ref{sec:SSKJ-WSD}, is an automatically constructed WSD dataset based on the sense inventory from the SSKJ dictionary. Lemmas with a single sense are discarded as they do not need WSD. In Section  \ref{sec:datasetEvaluation}, we
manually check if GPT-3.5 generates sensible and sense-aligned Slovene sentences from given dictionary definitions.

The second dataset, Elexis-WSD \cite{martelli2021designing}, is an existing dataset that contains 2024 sentences in 10 languages extracted from WikiMatrix \cite{schwenk2019wikimatrix}, a parallel corpus of Wikipedia sentences, and appended with manual translations for sentences missing from WikiMatrix. This dataset, described in Section \ref{sec:elexis-wsd}, is used to evaluate models' performance in WSD and WSI tasks. It is important to note that SSKJ-WSD and Elexis-WSD use different sense inventories and, therefore, cannot be combined into a single dataset.

Instead, we create a third dataset called SSKJ-Elexis-WiC, a word-in-context dataset built from pairs of sentences from SSKJ-WSD and Elexis-WSD, using only target lemmas from  Elexis-WSD. This dataset, described in Section \ref{sec:WiC}, is used to build WiC prediction models intended to solve not only WiC, but also WSD and WSI tasks. We present more details on the three datasets below. 

\subsection{SSKJ-WSD}
\label{sec:SSKJ-WSD}
%In order to construct a WSD dataset from dictionary definitions, we first limit the number of selected words. 
The Slovene SSKJ dictionary contains definitions for over 90,000 words, many of which rarely occur in practical use nowadays. Without loss of generality but to prevent excessive model sizes and training times, we limit the dataset only to lemmas that occur in the Elexis-WSD evaluation dataset, as those are relevant for the evaluation. This leaves us with 4,009 lemmas in total. Additionally, we ensure that every selected lemma contains more than one sense, as single-sense words would not be useful for evaluating WiC, WSD, and WSI tasks. Discarding lemmas with one sense leaves us 2,379 lemmas. Out of those, we limit our evaluation to the top 758 lemmas present in the Elexis-WSD dataset, as those outside the top 758 either contain a small number of examples or represent words that are not commonly used. For each lemma, we extract every usage example from the SSKJ dictionary and label it with the matching sense. This gives us the baseline SSKJ-WSD dataset consisting of 14,814 usage examples. 

As these usage examples are likely too short to be useful for the WSD task, we extend them using GPT-3.5. For each usage example, we generate 10 longer, complete sentences using the prompt "Razširi [zgled] v polno poved" (Expand [usage example] into a full sentence). This should produce 148,140 sentences in total, but sentences generated using GPT-3.5 sometimes contain errors that make them unusable. We automatically filter sentences that contain one of the two errors:
\begin{enumerate}
    \item The original dictionary lemma is not present in the full sentence. While we prompt GPT-3.5 to generate complete sentences by extending existing examples, it sometimes omits the original lemma.
    \item The generated sentence is identical to one of the already generated sentences. The 10 sentences generated by GPT-3.5 are not guaranteed to be unique; therefore, we discard duplicates.
\end{enumerate}
This leaves us with a total of 114,940 complete sentences describing 758 lemmas with 3,029 senses in total.

As described in Section \ref{sec:dictionary_structure}, the SSKJ dictionary contains its own sense inventory. Using this sense inventory and the extracted sentences, we construct the final SSKJ-WSD dataset, where every sentence is tagged with the corresponding SSKJ sense.

\subsection{Quality of the generated sentences}
\label{sec:datasetEvaluation}
%TODO-rewrite?
The core ideas of our work are i) to extract information from dictionaries by using LLMs to generate complete sentences from dictionary usage examples of a given lemma while preserving the original sense; ii) to use the generated sentences for training models for the WiC task, and iii) use WiC prediction models to solve also WSD and WSI tasks.
While we will automatically evaluate the sentence quality on downstream tasks in Section \ref{sec:results} (i.e., via performance on the WiC, WSD, and WSI tasks), we first manually verify the sense and quality of the GPT-3.5-generated sentences. If the generated sentences are grammatically incorrect, don't capture the correct sense of given words, or poorly match sentences produced by humans, they are unlikely to be useful for further tasks.

To evaluate these aspects, we first perform a manual analysis of the generated sentences. This evaluation is limited to sentence usability for the WiC task, while a complete analysis of Slovene sentences generated by GPT-3.5 is outside the scope of this work. Our analysis is quantitative, meaning that not every sentence needs to be correct, we just need enough correct sentences to benefit model performance.

With that in mind, we manually evaluated 200 randomly sampled sentences from the SSKJ-WSD dataset. In order for sentences to be useful for the WiC task, two conditions must hold true:
\begin{enumerate}
    \item The original dictionary word must be used in the same sense both in the dictionary usage example and the complete sentence. This is necessary to train supervised WiC or WSD models, as the training datasets need to be labeled. If the sense in the full sentence matches the sense in the original usage example, we can use the senses present in SSKJ as the labels in the dataset. Otherwise, the sentences would need to be manually labeled, removing the advantages gained by automatic sentence generation.
    \item The complete sentence extended from dictionary usage examples must be similar to sentences produced by humans. While LLMs are generally good at producing human-like text, it is possible that some of the generated sentences will be grammatically incorrect or nonsensical. 
\end{enumerate}

\begin{table}[htb]
\caption{\label{tab:word_results}An overview of the manual evaluation of complete sentences generated by the GPT-3.5 model. The table shows how many of the 200 generated sentences contain the word in the correct sense and how many contain human-like text.}

\centering
\begin{tabular}{lcc}
Condition & Number & \%\\
\hline
Correct sense & 199 & 99.5\\
Human like text & 197&  98.5\\ \hline
Both above & 197 & 98.5 \\
%Total sentences & 200 & 100\%\\

\end{tabular}
\end{table}

Table \ref{tab:word_results} presents the results, showing the number of sentences that pass each condition.
Despite the limited number of manually evaluated sentences, we get a basic picture of how well GPT-3.5 extends dictionary definitions into complete sentences. As a large proportion of the generated sentences (98.5\%) satisfies both conditions, GPT-3.5 will likely be useful for data augmentation when creating the Slovene WiC dataset.
%98.5\% of all examples generated sentences with human-like text that contained the word in its correct sense, with a longer sentence than the original example use-case. 

\begin{table*}[!!htb]
\caption{\label{tab:man_ana_correct}A sample of correct sentence expansions generated by GPT-3.5.
}
\centering
\begin{tabular}{p{0.25\linewidth}p{0.75\linewidth}}
Original snippet & Extended sentence \\
\hline
nihče ga ne pozna  & Nihče ga ne pozna, niti njegovi sosedje, niti njegovi kolegi, niti njegova družina. \\
nobody knows him & Nobody knows him, not even his neighbours, his colleagues or his family. \\ \hline
dva prsta širok rob & Rob je širok točno dva prsta. \\ 
two-inch-wide corner & The corner is exactly two inches wide \\ \hline
zvezna cesta Ljubljana–Celje  & Zvezna cesta Ljubljana–Celje je ena izmed najpomembnejših prometnih povezav v Sloveniji, ki povezuje glavno mesto Ljubljano s tretjim največjim slovenskim ... \\
The Ljubljana-Celje road & The Ljubljana-Celje road is one of the most important traffic connections in Slovenia, connecting the capital Ljubljana to the third largest slovene ...\\ \hline
zdravilo za srce & Zdravilo za srce je zdravilo, ki se uporablja za zdravljenje in preprečevanje bolezni srca ter ohranjanje zdravja srčno-žilnega
sistema.\\
heart medicine & heart medicine is a medicine used for treating and preventing hart disease and for maintaining the health of the cardiovascular system \\ \hline
blago prve vrste & Blago prve vrste je visokokakovostno blago, ki je običajno dražje od drugih vrst blaga. \\
first-grade goods & First-grade goods are high-quality goods and is usually more expensive than other kinds of goods \\
\hline
\end{tabular}
\end{table*}

Table \ref{tab:man_ana_correct} presents several correct sentences generated by GPT-3.5. Qualitative analysis shows that GPT-3.5 is capable of extending dictionary example use cases in multiple ways. Commonly, the use case is simply followed by its definition (heart medicine, first-class goods). Sometimes, the usage example is rearranged into a different form ("two-inches-wide-angle") that more commonly appears in general use. GPT-3.5 seems to contain factual knowledge of topics such as geography (the river Sava, Slovene roads) and medicine (heart medicine).

\begin{table*}[!!htb]
\caption{\label{tab:man_ana_incorrect}The two incorrect expanded sentences from the sample of 200 evaluated.}
\centering
\begin{tabular}{p{0.5\linewidth}p{0.5\linewidth}}
Original sentence & Extended sentence \\
\hline
veliki Z  & Veliki Z je bil izjemen voditelj in vizionar ... \\
Capital Z & Capital Z was a great leader and visionary ... \\ \hline
to žganje ima kakih trideset gradov  &  To žganje ima približno trideset stopinj Celzija.\\
This brandy has approximately 30 ABV & This brandy has 30 degrees Celcius \\
%peti ob spremljavi klavirja & Peti ob spremljavi klavirja je bila izjemno čustvena in ganljiva glasbena izvedba.\\
%Singing with piano accompaniment & Singing with piano accompaniment was an extremely emotional and touching musical performance \\
\hline
\end{tabular}
\end{table*}

Table \ref{tab:man_ana_incorrect} shows incorrect sentences generated by GPT-3.5. The first example shows how GPT-3.5 failed to generate correct sentences for one-letter words (e.g., the Slovene z, which can either mean the letter itself or can be used as a grammatical preposition). In the second example, GPT-3.5 fails to recognize technical terminology (i.e., alcohol percentage)
%, and in the third example, it mistakes a verb for a noun (i.e., mistakingly taking "Singing with piano accompaniment" as the title of a song). 
%\mrs{Zadnji stavek je pogojno pravilen, če gre za pripoved nekoga, ki je pel ob spremljavi klavirja? - \hl{odstranjeno}} 
Despite their relative rarity, incorrect sentences are likely to negatively impact the final performance of trained WiC models but are unlikely to outweigh the contributions made by the correct sentences.

\subsection{Elexis-WSD}
\label{sec:elexis-wsd}
Elexis-WSD is a multilingual WSD dataset containing parallel sentences from 10 languages (Bulgarian, Danish, Dutch, English, Estonian, Hungarian, Italian, Portuguese, Slovene, and Spanish). Each language contains 2024 sentences extracted from the WikiMatrix corpus. Parallel sentences missing from WikiMatrix were added with manual translation. Each word-sense is annotated using a sense inventory specific to each language. For Slovene, the Digital Dictionary Database of Slovene was used, which differs from the senses used by the SSKJ dictionary.

While both SSKJ-WSD and Elexis-WSD were created to solve the WSD task, we also use them when evaluating the WSI task.

\subsection{SSKJ-Elexis-WiC}
\label{sec:WiC}
%TODO double check, trenutno iz sskj podatkov
While both SSKJ-WSD and Elexis-WSD datasets contain sense annotations, they do not use the same sense inventory. This problem may be common to many languages, as existing dictionary and datasets may not use a unified sense inventory. To address this, we create two word-in-context datasets called SSKJ-WiC and Elexis-WiC. For each corresponding WSD dataset, we randomly sample pairs of sentences containing the same target lemma and then annotate the pair with 1 if target lemmas come from the same sense and 0 if the senses differ. We construct these datasets automatically. For each target lemma in a sentence, we pick 12 sentences, half with the same sense and the half with different senses, and then generate all possible sentence pairs. This gives us a balanced binary dataset with 758 target lemmas and 9,096 sentences in total.

While the WiC task simplifies the word sense disambiguation problem from a multi-class WSD task to a binary classification, it still requires the classification model to understand the correct sense of a word, allowing us to examine whether LLM-generated sentences contain such information. Importantly, the WiC task also removes the need for a sense inventory, allowing us to combine sentences from multiple datasets. We combine the sentences from Elexis-WiC and SSKJ-WiC into the SSKJ-Elexis-WiC dataset.

\section{Evaluation Scenarios and Results}
\label{sec:results}
In this section, we perform quantitative empirical evaluation and show that the generated datasets produce well-performing models and that the  GPT-generated sentences improve performance on WiC, WSD, and WSI tasks compared to unextended dictionary usage example snippets. We start by presenting the evaluation scenarios in \ref{sec:evaluation-scenarios}, followed by the results split into the three tasks.

\subsection{Evaluation scenarios}
\label{sec:evaluation-scenarios}
The analysis in Section \ref{sec:datasetEvaluation} shows that GPT-3.5 can generate Slovene sentences from dictionary use cases but does not show how these sentences affect WiC, WSD, and WSI performance. In order to evaluate this, we test our approach on three tasks:
\begin{enumerate}
    \item We use the generated sentences to train a \textbf{WiC} model. This approach does not require a pre-constructed sense dictionary.
    \item  We use the models trained for the WiC task to solve the \textbf{WSD} task. We compare models with extended sentences and models that do not use LLM-generated sentences.
    \item We evaluate the WiC models on the \textbf{WSI} task. This approach is similar to using WiC for the WSD task, but we simulate the induction of new senses when the target sense does not match any of the existing senses.
\end{enumerate}

We further split each task into three subtasks based on the overlap of target lemmas in the training and testing set. A unique advantage of our approach is that the SSKJ dictionary contains most of the words present in standard Slovene, something that is almost never the case with specific WSD datasets. This makes it particularly useful for out-of-vocabulary (OOV) WSD evaluation, where a (manually labeled) training set does not contain every word present in the test set. This type of evaluation is important for less-resourced languages, where manually labeled datasets often contain only a small subset of words used in the language. 

To specifically test our approach on OOV tasks, we split our evaluation into three subtasks:

\begin{enumerate}
    \item \textbf{Pure-OOV}, where the target words in the training set differ from those in the test set. In this task, we use part of the \emph{Elexis-WiC} as a test set and train the model on a subset of the \emph{SSKJ-WiC} dataset that excludes all target lemmas present in test set. 

    This subtask is the most difficult, as it does not fully use the data available in the SSKJ dictionary. A key advantage of using a dictionary as a source of training data is that it contains most standard words of a chosen language. However, constructing a WiC dataset that contains all words is not practical. Therefore, we use the Pure-OOV task to determine whether the sentences generated by GPT-3.5 contain enough information to be helpful in the disambiguation of words that do not appear in the training set.

    \item \textbf{Partial-OOV}, where the Elexis-WiC target lemmas differ in the training and test set, but the SSKJ-WiC dataset contains words from both the training and the testing set. In this task, we first split \emph{Elexis-WiC} into two subsets, each containing wholly distinct target lemmas. We then train the model on \emph{SSKJ-WiC} and one of the two \emph{Elexis-WiC} subsets, leaving the other \emph{Elexis-WiC} as a test set. 

    This subtask makes strong use of the LLM-generated sentences in the SSKJ-WiC dataset. It simulates a common real-world environment where a small manually-labeled dataset is available but does not contain every word we would like to disambiguate. In such a case, having access to a dataset constructed from a dictionary can help disambiguate words that are not present in the manually-labeled dataset.
    
    \item \textbf{Non-OOV}, where both the training and test sets contain the same target lemmas. This approach directly compares the quality of manually annotated sentences with the sentences generated by GPT-3.5. The evaluation process is similar to the Partial-OOV task, except that the two \emph{Elexis-WiC} subsets contain the same target subwords.

    This subtask makes the least use of the GPT-3.5 sentences since the training \emph{Elexis-WiC} subset already contains manually labeled examples of all target lemmas in the test set.  We use this subtask to examine whether GPT-3.5 sentences still contain some additional information that can be useful in Non-OOV scenarios.
\end{enumerate}

 We report the performance of our model on WiC, WSD and WSI tasks in Sections \ref{sec:WICevaluation}, \ref{sec:WSDevaluation} and \ref{sec:WSIevaluation}. For each task, we evaluate three types of splits: Pure-OOV, Partial-OOV and Non-OOV. 
 
As a baseline we also perform the same set of experiments using the \textbf{SSKJ-BASE-WSD} dataset (i.e. the WiC dataset created from the original SSKJ usage examples without the GPT-3.5-generated sentence expansions). This baseline tests how the contributions from GPT-3.5-expanded sentences compare with unmodified dictionary usage examples. 

To ensure manageable training times we limit the number of examples in all datasets. Since LLMs can generate many sentences and we obtain the final dataset by selecting pairs of sentences, we can generate a massive (quadratic) number of sentence pairs. While such large datasets may improve model performance, we limit the number of examples in each dataset so that training times remain reasonable. In the baseline SSKJ-BASE-WSD and GPT-3.5-extended SSKJ-WSD datasets, we limit each sense to the top six examples and 100 sentence pairs. Since the baseline SSKJ-WSD dataset rarely contains more than six examples and the examples generated by GPT-3.5 either degrade in quality or start repeating relatively early, this assures a good balance between dataset size and quality.

\subsection{WiC results}
\label{sec:WICevaluation}

To build the WiC classifier, we used the BERT-based SloBERTa model \cite{ulcar2021sloberta} and fine-tuned it for 20 epochs with the weight decay set to 0.01 and the AdamW optimizer with the learning rate set to $10^{-5}$. After 20 epochs, the accuracy on the training and validation sets stopped increasing. 

%\mrs{There seem to be some ambiguity here. Above we say that we use fine-tuning, below we speak of pretraining. Is this correct? I assumed that we only do fine-tuning. I also changed the names of some datasets. Please, check if the current description matches what was actually done. - \hl{Popravljeno v fine-tuning namesto pretraining}}
The results are presented in Table \ref{tab:final-wic-results}. When fine-tuning is not used, the model is trained and tested only on Elexis-WiC dataset, i.e. the WiC data obtained from the Elexis-WSD dataset. We compare these results with those obtained when performing additional fine-tuning on WiC examples obtained from SSKJ-BASE-WiC and SSKJ-WiC. It is clear that the default SSKJ word definitions do not provide additional information. On Pure-OOV and Partial-OOV tasks, the classifiers trained on SSKJ usage examples do not outperform the default classifier. On the Non-OOV task, the performance is the same as when fine-tuning is omitted. 

On the other hand, usage examples sentences generated by GPT-3.5 do contain useful information for training WiC models. Even on the Pure-OOV task, the model provides useful information, considerably above the default classifier, despite being trained purely on automatically generated sentences. On Partial-OOV and Non-OOV tasks, the GPT-3.5 sentences improve the performance, while dictionary snippets do not. 

\begin{table}[htb]
\caption{\label{tab:final-wic-results} The classification accuracy (CA) on the three WiC task types, using models with different information.}
\centering
\resizebox{\columnwidth}{!}{
\begin{tabular}{lccc}
Dataset used & Task type & CA & Default\\
\hline
ElexisWiC & Pure-OOV & 0.5 & 0.5 \\
ElexisWiC+Snippets & Pure-OOV & 0.5 & 0.5 \\
ElexisWiC+Sentences & Pure-OOV & \textbf{0.620} & 0.5 \\ \hline
ElexisWiC & Part-OOV & 0.731  & 0.639 \\
%Only GPT & Part-OOV & 0.673 (c)& 0.639 (c)\\
ElexisWiC+Snippets & Part-OOV & 0.731  & 0.639 \\
ElexisWiC+Sentences & Part-OOV & \textbf{0.742} & 0.639 \\ \hline
ElexisWiC & Non-OOV & 0.769  & 0.657 \\ 
ElexisWiC+Snippets & Non-OOV & 0.769  & 0.657 \\
ElexisWiC+Sentences & Non-OOV & \textbf{0.784} & 0.657\\
\hline
\end{tabular}
}
\end{table}

These results confirm that our approach of expanding dictionary usage examples into full sentences is sensible. We therefore continue with the WSD and WSI tasks.

%TODO maybe removes
%Table TODO shows the performance of each model for each epoch of training. Notably, pretraining the models on SSKJ-WSD shows significant improvements in accuracy during early epochs. This is consistent with the model performance on the OOV tasks, which showed a model trained solely on SSKJ-WSD is alredy able to predict sentences contained in Elexis-WSD. An important benefit of this is that the model needs less training to obtain equal results to a model trained solely on Elexis-WSD, which may be important in resource-limited environments or when training models with larger vocabularies.

\subsection{WSD evaluation}
\label{sec:WSDevaluation}
In addition to the WiC task, we evaluate the usability of dictionary usage examples and GPT-extended sentences on the more common WSD task. 

Since the sense inventories differ between the SSKJ dictionary and Elexis-WSD, we cannot simply train a WSD model on the SSKJ data and evaluate it on Elexis-WSD. Instead, we use the results of the WiC task to assign the senses. We pair each target sentence with all sense-labeled sentences in the training set belonging to the same target lemma. We then compute WiC scores for each sentence pair using the WiC models trained in Section \ref{sec:WICevaluation} and select the sense from the pair where the WiC score is highest. This allows us to assign senses based on the sense dictionary of the test set.

The results of the WSD evaluation are presented in Table \ref{tab:final-wsd-results}. Note that the Pure-OOV evaluation cannot be performed for the WSD task, as the task requires sense-labeled sentences of the same lemma to be present in the training set for the purpose of comparison. As with the WiC evaluation, expanding dictionary usage examples into sentences improves results for both Part-OOV and Non-OOV evaluations. The obtained CA scores (93\% for Part-OOV and 94.2\% for Non-OOV) are high enough to allow for practical use. The scores are also substantially higher than the scores without sentence expansion.

However, a direct comparison of our results with other approaches is difficult. Slovene currently does not have strong WSD or WSI models and multilingual approaches commonly under-perform due to the lack of available training data. Zero-shot approaches from English are possible using pre-trained multilingual word embeddings, but do not achieve good results \cite{bevilacqua2021recent}. In English, dedicated WSD models commonly have an upper performance bound caused by relatively low inter-annotator agreement (estimated to be around 80\% for fine-grained sense inventories \cite{navigli2009word}. We avoid this disagreement in our training set, as we construct it from dictionary definitions without manual annotation. The Elexis-WSD dataset, which we use for testing, was constructed by experts and we can assume perfect annotations. This makes our results look excellent. However, the Slovene subset of Elexis-WSD contains a smaller set of senses than datasets used for benchmarking in English, making direct comparison to English benchmarks difficult. 

%In fact, for the less-resourced Slovenian language we obtain results results on par or better than the reported results for English.

%\mrs{Te rezultate bi bilo vseeno smiselno primerjati z objavljenimi rezultati za druge jezike.}
%\mrs{Please check if the last claim is true in the Elexis-WSD paper or Navigli's group work on WSD. - \hl{ne najdem nobenih res primerljvih rezultatov, tako da sem to trenutno odstranil}}

%TODO - double check sskj-pretraining results
\begin{table}[htb]
\caption{\label{tab:final-wsd-results}. The classification accuracy (CA) on the two WSD task types, using models with different information.}
\centering
\resizebox{\columnwidth}{!}{
\begin{tabular}{lccc}
Model & Task & CA & Default\\
\hline

ElexisWiC & Part-OOV & 0.868 & 0.728 \\
ElexisWiC+Snippets & Part-OOV & 0.868  & 0.728 \\
ElexisWiC+Sentences & Part-OOV & \textbf{0.930} & 0.728 \\ \hline

ElexisWiC & Non-OOV & 0.922  & 0.728 \\ 
ElexisWiC+Snippets & Non-OOV & 0.922 & 0.728 \\
ElexisWiC+Sentences & Non-OOV & \textbf{0.942} & 0.728\\
\hline
\end{tabular}
}
\end{table}

\subsection{WSI evaluation}
\label{sec:WSIevaluation}
We also extend our approach to WSI. The methodology is similar to the WSD task with one modification: if the WiC scores of all sentence pairs are below a certain threshold, we induce a new sense and add it to the sense inventory. We set the threshold by first performing the evaluation on a validation set (10\% of the Elexis-WSD not included in the training and testing sets) and selecting the best-performing threshold. In all cases, this was obtained when the probability for no sense was higher than 1.2 times the average prediction probability of all predictions made by the neural network on the validation set. %(i.e., 2.976). 
%\mrs{The mentioned scores are not clear. Is 1.2 a percentage? Can you explain better what we measure? \hl{It's the average value * 1.12}} 
As with WSD, OOV evaluation is not possible during this task.

The results of the WSI evaluation are presented in Table \ref{tab:final-wsi-results}. Again, using no additional dictionary data or only usage example snippets provides no improvement, while using GPT-generated sentences improves the results in the Part-OOV and Non-OOV evaluation types. The obtained CA scores (75.1\% for Non-OOV and 70\% for Part-OOV) are high enough to allow for practical use, i.e. detection of new senses for words existing in dictionaries with human supervision. %\mrs{Large difference between Part-OOV and Non-OOV is suspicious. Can we explain it? \hl{Tole sem še enkrat pognal. Sedaj so rezultati slabši, ampak bolj logični (Part-OOV slabše kot OOV},}.

\begin{table}[htb]
\caption{\label{tab:final-wsi-results}. The classification accuracy (CA) on the two WSI task types, using models with different information.}
\centering
\resizebox{\columnwidth}{!}{
\begin{tabular}{lccc}
Model & Task & CA & Default\\
\hline
%ElexisWiC & Part-OOV & 0.821 & 0.506 \\
%ElexisWiC+Snippets & Part-OOV & 0.821  & 0.506 \\
%ElexisWiC+Sentences & Part-OOV &  \textbf{0.891} & 0.506 \\ \hline
ElexisWiC & Part-OOV & 0.661 & 0.506 \\
ElexisWiC+Snippets & Part-OOV & 0.661  & 0.506 \\
ElexisWiC+Sentences & Part-OOV &  \textbf{0.700} & 0.506 \\ 
\hline
ElexisWiC & Non-OOV & 0.721 & 0.510 \\
ElexisWiC+Snippets & Non-OOV & 0.721 & 0.510 \\
ElexisWiC+Sentences & Non-OOV & \textbf{0.751} & 0.510 \\ \hline
\end{tabular}
}
\end{table}

\section{Conclusions}
\label{sec:conclusions}
We presented a novel approach using the WiC task for WSD and WSI tasks. We show that generative LLMs can be used to expand snippets in dictionary usage examples into complete sentences that contain enough word-sense information to be useful when training WiC models. Our use case on the less-resourced Slovenian language shows that Slovene sentences generated by the LLM are semantically and syntactically correct and mostly reflect the correct sense of the target word. While the constructed datasets are not as good as datasets manually constructed by humans, the obtained sentences show potential for training machine learning models in scenarios where manually constructed datasets are small or do not exist. 

We intend to test the presented WiC-based methodology for WSD and WSi in the automatic construction of dictionary senses in the context of lexicography. This might require the creation of larger training sets. Further use is expected in the area of diachronic linguistic analyses where the WiC model might be able to detect when a new sense of a word emerges.
%\mrs{We saw no human-generated datasets for comparison. Can we provide such information from other languages?}

%TODO double check 9096
Although our work shows promising results, it provides only a preliminary look into linguistic dataset generation using Slovene LLMs. 
Data generation using LLMs allows for the generation of extremely large datasets. Due to time and resource constraints, we limited the size of our data to 9,096 sentence pairs. While we obtained promising results using such a dataset, we could create datasets many times larger. How larger datasets would benefit this task is a topic for further investigation. LLMs could improve performance on several NLP tasks in other less-resourced languages.

Qualitative analysis of sentences generated by GPT-3.5 and other LLMs for many less-resourced languages is still largely unexplored. Our work only examined a small number of Slovene sentences to determine whether they were useful for the WiC task. Further analysis is needed to obtain a deeper understanding of how different LLMs generates Slovene sentences and on which types of senses they fail. 

\section*{Acknowledgements}
%Skipped due to anonymization. To be included in the final version.
The work was partially supported by the Slovene Research and Innovation Agency (ARIS) core research programme P6-0411, as well as projects CRP V5-2297, L2-50070, and GC-0002. The work was also supported by EU through ERA Chair grant no. 101186647 (AI4DH) and cofinancing for research innovation projects in support of green transition and digitalisation (project PoVeJMo, no. C3.K8.IB).

%\mrs{The references need some cleaning, e.g., ArXiv replacement with journals and conference versions, capitalization \hl{Poprabljeno kolikor se je dalo, nekateri članki imajo samo arxiv}}

%\clearpage
%\section*{Acknowledgements}
%Skipped due to anonymization. To be included in the final version.
%The work was partially supported by the Slovene Research and Innovation Agency (ARIS) core research programme P6-0411, as well as projects CRP V5-2297, L2-50070, and GC-0002. The work was also supported by EU through ERA Chair grant no. 101186647 (AI4DH) and cofinancing for research innovation projects in support of green transition and digitalisation (project PoVeJMo, no. C3.K8.IB).

\nocite{*}
\section{Bibliographical References}\label{sec:reference}

\bibliographystyle{lrec-coling2024-natbib}
\bibliography{lrec-coling2024-example}

%\section{Language Resource References}
%\label{lr:ref}
%\bibliographystylelanguageresource{lrec-coling2024-natbib}
%\bibliographylanguageresource{languageresource}

\end{document}